\pdfoutput=1
\documentclass[11pt]{article}
\usepackage{colortbl}
\usepackage{color}
\usepackage[dvipsnames]{xcolor}
\definecolor{Greenish}{rgb}{0.10,0.60,0.30}

\usepackage[preprint]{acl}

\usepackage{times}
\usepackage{latexsym}

\usepackage[T1]{fontenc}
\usepackage[utf8]{inputenc}

\usepackage{microtype}

\usepackage{inconsolata}
\usepackage{enumitem}

\usepackage{graphicx}

\title{Quantifying the Gaps Between Translation and Native Perception in \\Training for Multimodal, Multilingual Retrieval}

\author{Kyle Buettner\textsuperscript{\rm 1},
    Adriana Kovashka\textsuperscript{\rm 1,2}\\
\textsuperscript{1}Intelligent Systems Program, \textsuperscript{2}Department of Computer Science,
University of Pittsburgh \\
{\tt buettnerk@pitt.edu, kovashka@cs.pitt.edu}\\
\small \href{https://krbuettner.github.io/MultilingualRetrievalStudy}{https://krbuettner.github.io/MultilingualRetrievalStudy}
}

\usepackage{colortbl}
\usepackage{color}
\usepackage[dvipsnames]{xcolor}
\definecolor{Gray}{gray}{0.77}
\definecolor{DarkGray}{gray}{0.43}
\definecolor{LLGray}{gray}{0.91}
\definecolor{LightCyan}{rgb}{0.88,1,1}
\definecolor{Greenish}{rgb}{0.10,0.60,0.30}
\definecolor{ForestGreen}{RGB}{34,139,34}
\definecolor{Wood}{RGB}{150,111,51}
\definecolor{DarkBrown}{RGB}{150,78,2}
\definecolor{Pinkish}{RGB}{255,102,220}
\definecolor{Blueish}{RGB}{2,78,150}

\newcolumntype{a}{>{\columncolor{Gray}}c}
\definecolor{LGray}{gray}{0.94}

\newcolumntype{g}{>{\columncolor{LGray}}c}

\begin{document}
\maketitle
\begin{abstract}

    There is a scarcity of multilingual vision-language models that properly account for the perceptual differences that are reflected in image captions across languages and cultures. In this work, through a multimodal, multilingual retrieval case study, we quantify the existing lack of model flexibility. We empirically show performance gaps between training on captions that come from native German perception and captions that have been either machine-translated or human-translated from English into German. To address these gaps, we further propose and evaluate caption augmentation strategies. While we achieve mean recall improvements (+1.3), gaps still remain, indicating an open area of future work for the community.

\end{abstract}

\section{Introduction}
\label{sec:intro}

    Vision-language models (VLMs)  such as CLIP \citep{radford2021learning} are predominantly limited to use in English as a result of the pretraining supervision consisting mostly of English captions. This trend naturally poses an accessibility barrier for non-English speakers. Furthermore, cultures around the world differ in their salient concepts \citep{liu2021visually} and visual perception \cite{nisbett2013culture}.
    Relying on English supervision in pretraining thus hinders consideration of cross-cultural concepts in object-based tasks such as recognition, detection, and image-text retrieval. 

\begin{figure}[t]
        \centering
        \includegraphics[scale=0.34]
        {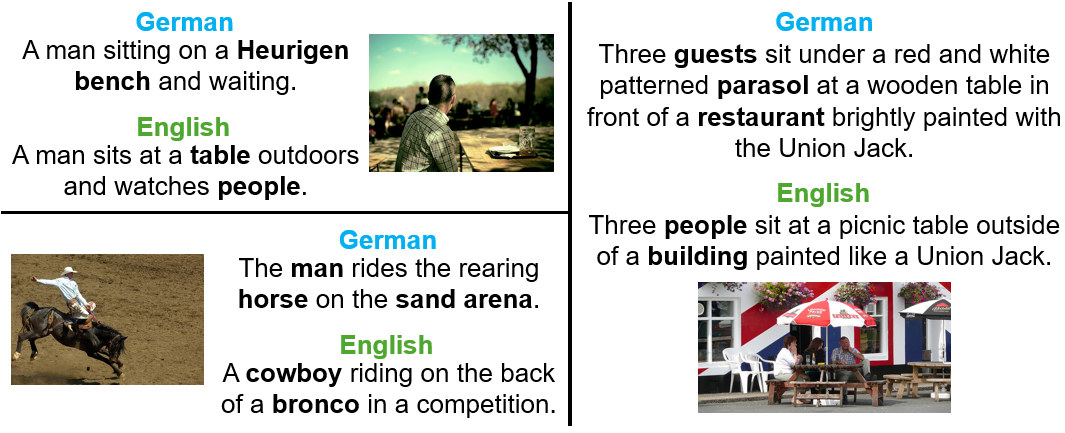}
        \caption{\textbf{Example perception differences between native English and German speakers.} Examples are captions from Flickr30K \cite{young-etal-2014-image} and Multi30K \cite{elliott-etal-2016-multi30k}. Note differences in mentioned objects (``sand arena'', ``parasol'') and specificity (``Heurigen bench'' vs. ``table'', ``horse'' vs. ``bronco''). German captions here are translated to English.}
        \label{intro_fig}
    \end{figure}

    Example cultural differences present in language with respect to object \textit{specificity} and \textit{importance}. For example, past literature \citep{nisbett2013culture} describes differences in how cultures perceive members of an object group (e.g. penguins within the group of birds), indicating that certain groups have stronger associations for \emph{specific} rather than general object terms. Experiments in \citet{nisbett2013culture} also illustrate differences between East Asians and Americans with respect to the perceived \emph{importance} of background objects and context as opposed to foreground objects. Different cultures notice different objects more; perceptual differences may manifest in objects being included/excluded in a caption, and different objects being relevant in tasks. Fig. \ref{intro_fig} shows examples of differences in AI datasets for English and German \citep{young-etal-2014-image, elliott-etal-2016-multi30k}.

    There has been some progress in multilingual, multimodal modeling \citep{chen2022pali, chen2023pali, carlsson2022cross-m-clip, chen-etal-2023-mclip} and multilingual data creation \citep{elliott-etal-2016-multi30k, yoshikawa-etal-2017-stair, liu2021visually, ThapliyalCrossmodal2022}. The models often leverage off-the-shelf machine translation techniques to improve multilingual functionality. In this work, we investigate the performance gaps between training models with translations (which reflect English speaker perception) and natively written captions (which reflect non-English speaker perception) for a task in a given language. In line with the observed differences in \citet{nisbett2013culture}, we reason that translation may not account for specificity differences and may not alter supervision to account for importance differences.
    
    %If included in the captions, certain objects can function as distractors, leading to undesirable and/or unnecessary correlations being learned for a given language/culture. 

    We quantify potential differences through an exploration of non-English image-text retrieval. In particular, we finetune and benchmark multilingual CLIP \cite{chen-etal-2023-mclip} on Multi30K \cite{elliott-etal-2016-multi30k} using \textit{German} as the target. We explore Multi30K's native German captions (reflecting German speaker perception) and professionally translated captions (from English to German), as well as use of an external machine translation model over Flickr30K's English captions \cite{young-etal-2014-image}. We find significant performance differences depending on the data used to train the model, i.e. (1) English, (2) German translated from English by a machine translation model, (3) German translated from English by humans, and (4) native German. 

    As (2) and (3) have gaps vs. (4), we also attempt to improve upon  translation. We test three paraphrasing techniques to diversify object descriptions in English before translation, and use the resulting translations as additional finetuning data. \textit{First}, we experiment with a \textit{hypernymization} data augmentation technique, where object terms are updated before translation to represent different levels of specificity. \textit{Second}, we use a large language model (LLM), LLaMA-3 \citep{touvron2023llama}, to produce \textit{structurally different, but semantically similar} paraphrases of English captions before translation. \textit{Third}, we explore LLM reasoning to produce \textit{targeted} paraphrases that capture the perceptual properties captured in a sample set of captions. These techniques outperform the baselines. \textbf{However, a gap between translation and native perception remains, indicating an open problem.} We conclude with analysis in pursuit of this direction.

\section{Background and Related Work}

    \noindent \textbf{Cultural differences in perception.}
    Prior work considers how culture may influence perception and expression. For example, Western and East Asian cultural differences are found to manifest in visual attention, e.g. Americans appear to pay more attention to foreground/objects than East Asians, but conversely for background/context \citep{nisbett2013culture}. Furthermore, \citet{boroditsky2006linguistic} describes empirical studies that indicate that different cultures group objects differently (e.g. based on shape or material) and ascribe different properties to objects, because of unique grammar (e.g. gendered nouns). Since German uses gendered nouns, this observation may manifest in native German captions (and retrieval) as objects being described with unique attributes.  
    The work of \citet{berthele2015static} notes that Germanic language speakers describe object relationships with notably specific spatial information (e.g. posture/manner information in addition to object relationships). \citet{hofstede2001culture} conducts analysis to show that there are cultural differences between Germany and United States in terms of individualism vs. collectivism, which could impact the perception of visual content as argued by \citet{nisbett2013culture}.
    Further examples can be found in work on linguistic relativity \citep{kay1984sapir}.

    \noindent \textbf{Multilingual multimodal modeling.} Our work aligns with works that extend VLMs for use in languages besides English \citep{chen2022pali, chen2023pali, carlsson2022cross-m-clip, chen-etal-2023-mclip}. Models notably often rely on translations, and works do not have analysis into performance differences between translations and captions of native perception. In contrast, \citet{kadar2018lessons} and our work show differences in retrieval performance when captions are natively written in a language or translated into that language from English. Our work differs from \citet{kadar2018lessons} as we also explore machine translation, with a more modern VLM \citep{chen-etal-2023-mclip}. We also explicitly address the lack of techniques to overcome gaps by experimenting with paraphrasing augmentations. Our strategies are related to past paraphrasing work \cite{wieting-gimpel-2018-paranmt, hu2019parabank}, but these approaches use machine translation to generate large-scale English paraphrase datasets, while we leverage in-context learning and LLMs to generate paraphrases \textit{for use as input to machine translation to enhance diversity}. We are inspired by \citet{fan2024improving}, as the work shows zero-shot image classification improvements with LLM-based caption rewriting. 

    Data-wise, we explore Multi30K \citep{elliott-etal-2016-multi30k} as it contains native German captions and parallels the English Flickr30K captions \citep{young-etal-2014-image}. XM3600 \citep{ThapliyalCrossmodal2022} also provides natively perceived captions, in 36 languages for 3,600 images. Due to size, we do not train with this set, though we provide initial analysis on it to inspire future work. WebLI \cite{chen2022pali} is another dataset that contains crawled captions in 109 languages, though it is proprietary.

% While this setting is realistic, the work of \citet{kadar2018lessons} shows benefits in performance when techniques can be employed with captions in multiple languages \emph{for the same image}.
     
    %We in general aim to cope with these found differences.   in the lack of generalization through study of techniques in an effort 

% , have different statistics,
    
  %   \citet{chen2022pali} achieve cross-language ability through a diverse mixture of training tasks and \citet{chen-etal-2023-mclip,carlsson2022cross-m-clip} through multilingual embeddings and machine translation. However, none    

\section{Experimental Methodology}

    We benchmark training with captions that reflect native perception by German speakers, ones that have been \textit{machine-translated} from captions reflecting English speaker perception, and ones \textit{human-translated} from English speaker perception. We also test strategies to improve upon translation.

% We aim to quantify performance differences in multimodal, non-English retrieval when training with captions coming from the perception of native and non-native speakers. 

\subsection{Benchmarking Details}

\noindent \textbf{Task.} We evaluate on German image-text (I2T) and text-image (T2I) retrieval. The German captions used in eval are written directly by native speakers about images. They are \textit{not} translated from English and represent natural non-English perception.  

\noindent \textbf{Data.} English data is from Flickr30K \cite{young-etal-2014-image}, and German data is from Multi30K \citep{elliott-etal-2016-multi30k}. Flickr30K contains 31,014 images that are annotated with 5 independently written English captions per image. Likewise, Multi30K provides 5 independently written German captions for the same images. These German captions are collected from 185 native speakers using a similar interface to Flickr \cite{hodosh2013framing}. Multi30K also provides professional German translations. In particular, for each image, 1 of the 5 English captions is sampled from Flickr30K, and professional translators produce corresponding captions in German (just from source text, not using the images). We refer to the separate caption sets as \textit{Independently Written} (5 sets for each language) and \textit{Human-Translated} (1 set per language). For all sets, we randomly split data to create a disjoint reference set (9,666 samples) to be used with our strategies (Sec.~\ref{sec:methods_compared}), as well as retrieval train/val/test sets (9,666/1,014/10,668 samples respectively).

\noindent \textbf{Modeling.} We explore mCLIP
%\textit{m}-CLIP 
\citep{chen-etal-2023-mclip}, an approach which has made CLIP multilingual through knowledge distillation-based training of projector modules and replacement of CLIP's text encoder with the multilingual text encoder XLM-R \citep{conneau-etal-2020-unsupervised}. We \textit{finetune} mCLIP with images and captions for 
German I2T and T2I retrieval. 
For experimentation that involves machine-translating English captions to German, we use \textit{opus-mt-en-de} 
\citep{TiedemannThottingal:EAMT2020} from Hugging Face. With this model, we use a deterministic setting, where tokens are generated according to highest token probability, and infer at most 40 tokens for each caption. mCLIP models are trained for 30 epochs on 1 Quadro RTX 5000 GPU with batch size 16 and learning rate 0.0005. 

\noindent \textbf{Metric.} We report \textit{mean recall} as in \citet{chen-etal-2023-mclip}. Recall@1,5,10 is computed for both T2I and I2T retrieval on each native German test set (5 sets total).
%(leading to 6 total values). 
\emph{Mean recall} is the average of these six values. We further average over each set.

\subsection{Methods Compared}
\label{sec:methods_compared}

\noindent \textbf{Baseline} finetuning strategies include: 
%We first test \textbf{baselines and upper bounds} to establish task difficulty with mCLIP:

\textbullet \space \textsc{Eng}, a ``lower bound'': finetuning using data natively provided in English (in the \textit{Independently Written} sets). Since there are 5 sets of captions, %performance in this setting is the 
we average over trials using each set for training. 
%This is a lower bound (since not in German).

\textbullet \space\textsc{Eng2Ger-MT:} finetuning on German sentences that have been translated from English using an English-to-German machine translation model \citep{TiedemannThottingal:EAMT2020}. English sentences come from the \textit{Human Translation} set.

\textbullet \space \textsc{Eng2Ger-MT (Trn):} same as above, but the translation model is further trained on captions from the Multi30K disjoint reference split we create, with the intuition that translation finetuning may capture caption differences. 
We train for 10 epochs with learning rate 0.00001 and batch size 16, using the \textit{Human Translation} pairs.
%For translation training, we use \textit{Human Translation} parallel pairs to gauge feasibility of training. Retrieval is finetuned using translated English data, though using native sets since the \textit{Human Translation} pairs are used for translation training. Performance is the average over trials using each native set for training. 

%The intuition is that captions require their own specific MT algorithm
%due to their being a representation of the visual world

%\noindent \textbf{Upper bounds} include:
\textbullet \space \textsc{Eng2Ger-HT}: finetuning on German captions translated from English by professional annotators (in the \textit{Human Translation} set). This training is different from and expected to perform worse than native German, but better than naive translation.

\textbullet \space \textsc{Ger}: finetuning using data natively provided from German perception (in the \textit{Independently Written} sets). Since there are 5 sets of captions, %performance in this setting is the 
we average over trials using each set for training. 
%Since there are 5 sets of captions, performance in this setting is the average over using each set for training. This is an upper bound (native German is used).

\noindent \textbf{Strategies:} We find significant gaps between these methods, notably \textsc{Eng2Ger-MT} and \textsc{Ger}, motivating experimentation with potential improvements. We test adding training data that has been augmented in English then translated to German. Some proposed changes involve object names, so for this purpose, we define an object vocabulary $\mathcal{V}$ including COCO object terms \citep{lin2014microsoft}. Category detection involves consideration of these terms, synonyms \citep{lu2018neural}, plurals, and word sense. For each strategy, mCLIP is trained as in \textsc{Eng2Ger-MT}, but with an augmented dataset of captions added. Methods include:

\textbullet \space \textsc{Hyper:} After identifying each COCO class with a synset id, if available, object mentions are hypernymized to be a random term above it in the WordNet hierarchy \citep{miller1995wordnet}. %Specifically, we perform a closure up the WordNet hierarchy for the possible hypernyms of a category. 
Our goal is to improve robustness to changes in object naming to address challenges in object specificity.  
 
%[Q to Kyle: ``all levels up'' -- presumably not to ``entity'' but to some shared hypernym? Probably some shared level is the top level?] 
 
%Add the hypernymized captions to the original training data, and fine-tune as above. The intuition for this approach is due to the differences in English and German, where German uses more specific terms.

%[Q to Kyle: but intuition seems backwards, i.e. hyponymizing should help, not hypernimizing, because German is more specific. Other intuition for the success of this? Just diversifying the data? Less in sync with our narrative...]

 \textbullet \space \textsc{Para-Rnd} (paraphrase-random): Before translation, we ask LLaMA-3 \citep{touvron2023llama} to write each caption in a structurally different manner while maintaining meaning. We are motivated by \citet{fan2024improving} which shows English retrieval benefits from diversification. Our approach differs as we diversify before translation to guide translation to more generalizable descriptions.  %Please refer to supp. for the specific prompts and configuration.

 \textbullet \space \textsc{Para-Tgt} (paraphrase-targeted): We ask LLaMA-3  to paraphrase each caption using examples of object naming ``style''. For each caption, a total of $k$=100 captions are randomly sampled from the reference split of the first native German set, such that if possible, sampled captions share at least one non-person object mention with the current caption (since most captions mention people). Translations of these are provided in the LLaMA-3 prompt as examples. Then for the input caption, LLaMA-3 is instructed to find relevant noun phrases, and to convert the noun phrases to more aligned representations based on the examples. 

\textbullet \space \textsc{Para-Cmb} %(paraphrase-combined): 
combines both sets above.
%two methods are added to training. 

\noindent Please refer to the appendix for prompt details.

% Details to add above: Identifying object mentions, Text augmentation with LLMs and WordNet/SpaCy

\section{Key Findings}

    In the top block of Table \ref{tab:main}, zero-shot mCLIP is shown to achieve the lowest recall (24.5). Finetuning mCLIP with English Multi30K data improves performance to 26.9 (+2.4). English data can help to a degree on German retrieval due to alignment learned in pretraining the multilingual text encoder. However, much more significant gains are achieved when the finetuning data is in German. Training with German data that has been translated from English using an off-the-shelf translation model (\textsc{Eng2Ger-MT}) reaches 33.4 (second block). Compared to human translation (\textsc{Eng2Ger-HT} - fourth block), there is a notable gap from machine translation (3.4), and finetuning the translation model only bridges this gap by 0.6. These results indicate existing challenges with off-the-shelf translation for retrieval. Then most significantly, the gap between off-the-shelf translation and native German captions (\textsc{Ger}) is 5.0. There is a notable gap between professional translation (\textsc{Eng2Ger-HT}) and \textsc{Ger} (1.6), which we reason is the gap due to differences in English and German perception. For example, these gaps could be due to specificity and importance differences. Expert translation does \textit{not} address these factors. 
    
    In the third block, our methods are found to be somewhat effective for bridging the gap between \textsc{Eng2Ger-MT} and \textsc{Ger}. \textsc{Hyper} improves the result by 0.3, and \textsc{Para-Rnd} and \textsc{Para-Tgt} by 0.7. These models are notably more appropriate for low-resource target languages than \textsc{Eng2Ger-MT (Trn)} since they use no/few reference captions compared to translation finetuning. Further combining random and targeted paraphrasing results in the largest gain of 1.3. The result is still 3.7 away from \textsc{Ger}. Addressing differences in the perception of the visual world and the way captions are written across cultures is thus an open challenge.

          \setlength{\tabcolsep}{1.0pt}

\begin{table}[t]
    \centering
    \renewcommand{\arraystretch}{0.85}
    %\resizebox{\linewidth}{!}{
    \begin{tabular}{|c|c|c|}
    \hline
    \small \textbf{Method} & \small \textbf{Mean Recall} & \textbf{\small Vs. \textsc{Eng2Ger-MT}} \\
    \hline
    \hline
    \small \textsc{mCLIP} & 24.5 & \color{red}{-8.9} \\
    \small \textsc{Eng} & 26.9 & \color{red}{-6.5} \\
    \hline
    \small \textsc{Eng2Ger-MT} & 33.4 & 0.0 \\
    \small \textsc{Eng2Ger-MT (Trn)} & 34.0 & \color{Greenish}{+0.6} \\
    \hline
    \small \textsc{Hyper} & 33.7 & \color{Greenish}{+0.3} \\
    \small \textsc{Para-Rnd} & 34.1 & \color{Greenish}{+0.7} \\
    \small \textsc{Para-Tgt} & 34.1 & \color{Greenish}{+0.7} \\
    \small \textsc{Para-Cmb} & 34.7 & \color{Greenish}{+1.3} \\
    \hline
    \small \textsc{Eng2Ger-HT} & 36.8 & \color{Greenish}{+3.4} \\
    \small \textsc{Ger} & 38.4 & \color{Greenish}{+5.0} \\

    \hline
    \end{tabular}
    %}
    \caption{\textbf{German I2T/T2I retrieval results.} Mean recall values are averaged over native German cap sets.}
    \label{tab:main}
\end{table}

\section{Further Analysis}

\begin{table*}
\centering
\renewcommand{\arraystretch}{0.75}

{
\begin{tabular}{|c |c |g |c |g |c | g |c |g |c |g |c |}
\hline
 & \small \textbf{eu-mean} & \small \textbf{eu-stdev} & \small\textbf{ar-mean} & \small\textbf{ar-stdev} & \small \textbf{hi-mean} & \small \textbf{hi-stdev} & \small \textbf{id-mean} & \small \textbf{id-stdev} & \small \textbf{easia-mean} & \small \textbf{easia-stdev} & \small \textbf{sw}  \\ \hline
    \small \textbf{tree(s)} & \small 270.5$^{-}$ & \small 92.9 & \small 349 & \small 19.8 & \small \textbf{581.5}$^{+}$ & \small 214.3 & \small 286 & \small 49.5 & \small 274.7 & \small 63.5 & \small \textbf{383} \\ \hline
    
        \small \textbf{mountain(s)} & \small 171.1$^{-}$ & \small 47.8 & \small 183 & \small 24.0 & \small 185.5 & \small 31.8 & \small 173 & \small 42.4 & \small \textbf{218}$^{+}$ & \small 16.5 & \small \textbf{208} \\ \hline
    
        \small \textbf{street} & \small \textbf{100.9} & \small  30.2 & \small \textbf{124}$^{+}$ & \small 50.9 & \small 61 & \small 7.1 & \small 38.5$^{-}$ & \small 10.6 & \small 76.7 & \small  19.0 & \small 82 \\ \hline
    
        \small \textbf{car(s)} & \small 207.3 & \small 20.0 & \small 235 & \small 24.0 & \small \textbf{239} & \small 50.9 & \small 204$^{-}$ & \small 11.3 & \small 220 & \small 17.8 & \small \textbf{270}$^{+}$ \\ \hline
    
        \small \textbf{building(s)} & \small 244.8$^{-}$ & \small 69.3 & \small 281.5 & \small 40.3 & \small 329 & \small 108.9 & \small \textbf{383.5} & \small 84.2 & \small 253.3 & \small 49.9 & \small \textbf{502}$^{+}$ \\ \hline
    
      \small   \textbf{restaurant} & \small 45.8 & \small 13.7 & \small \textbf{54} & \small 7.0 & \small 19$^{-}$ & \small 5.7 & \small \textbf{50.5}$^{+}$ & \small 13.4 & \small 42.7 & \small 6.1 & \small 21 \\ \hline
    
        \small \textbf{table} & \small 156.7 & \small 52.8 & \small 162.5 & \small 58.7 & \small \textbf{240}$^{+}$ & \small 12.7 & \small \textbf{228} & \small 93.3 & \small 185.3 & \small 43.7 & \small 121$^{-}$ \\ \hline
    
       \small  \textbf{plate} & \small 112.5 & \small 25.9 & \small 90$^{-}$ & \small 12.7 & \small 105.5 & \small 10.6 & \small 109.5 & \small 33.2 & \small \textbf{119.3}$^{+}$ & \small 5.1 & \small \textbf{113} \\ \hline
    
        \small \textbf{box} & \small 18.1 & \small 4.5 & \small 15.5$^{-}$ & \small 0.7 & \small 15.5 & \small 2.1 & \small \textbf{28}$^{+}$ & \small 4.2 & \small \textbf{24} & \small 2.6 & \small 18 \\ \hline
    
        \small \textbf{bottle} & \small 10.2$^{-}$ & \small 2.7 & \small 12 & \small 0 & \small 10.5 & \small 2.1 & \small 11 & \small 4.2 & \small \textbf{14.7}$^{+}$ & \small 0.6 & \small \textbf{18} \\ \hline
    
       \small \textbf{dog} & \small 26.2 & \small 5.1 & \small 28 & \small 1.4 & \small \textbf{31.5}$^{+}$ & \small 5.0 & \small 29.5 & \small 0.7 & \small 20.7$^{-}$ & \small 5.5 & \small \textbf{34}  \\ \hline
    
       \small \textbf{woman} & \small 135.5 & \small 23.7 & \small 127 & \small 5.7 & \small 114$^{-}$ & \small 31.1 & \small \textbf{164.5}$^{+}$ & \small 20.5 & \small 133.3 & \small 27.7 & \small \textbf{160} \\ \hline
\end{tabular}
}
\caption{\textbf{Language shifts in terms of concept mentions in different languages.} We group XM3600 European languages (eu), Arabic/Farsi (ar), Hindi/Bengali (hi), Indonesian/Thai (id), East Asian languages (easia), and report Swahili on its own (sw). The largest two numbers per row are bolded. Observe the differences between the language with highest (+) and lowest (\texttt{---}) counts, which are significantly larger than the within-group standard deviations. }
%Note that standard deviations even within language groups are high.}
\label{tab:lang_stats}
\end{table*}

    \textbf{Object mentions in English/German captions.} To analyze possible differences in perception, we analyze object mention frequency in Flickr30K/Multi30K. We specifically translate German captions to English and extract nouns in both (original) English and (translated to English) German captions. The ratio of English and German mentions is about 1.5, i.e. English mentions object nouns 50\% more often than German. However, counts vary by object type. For example, English mentions clothing more often (pants-143\% more, shirt-112\%, hat-60\%, jacket-43\%), and German mentions furniture more often (table-37\% more, bed-20\%, bench-15\%). These languages also vary in granularity: English captions often say ``people'', while German ones say ``workers'', ``athletes'', etc. 

    \noindent \textbf{Analysis of other languages.} We conduct initial analysis of the languages and captions in XM3600 \cite{ThapliyalCrossmodal2022}. We group XM3600 languages into European, Arabic/Farsi, Hindi/Bengali, Indonesian/Thai, East Asian, and Swahili categories. After translating each language to English, we report average  mention counts and standard deviations per group for various common objects in Table \ref{tab:lang_stats}. Language groups show large differences in terms of how commonly they mention elements of nature (e.g. mountains, trees), scenery (streets, buildings), household objects (table, plate, box, bottle), and the gender of portrayed people. It is also found that the difference between objects counts across languages is much greater than within-group standard deviations. Such results suggest differences in supervision worthy of exploration.

    \noindent \textbf{Paraphrasing.} %In Sec.~\ref{sec:intro}, we observe some common patterns in German captions (and respectively, German captions literally translated into English). For example, German mentions specific categories/entities of people, e.g. the entity ``bicyclist'' instead of ``person riding a bicycle''. 
    LLaMA picks up on granularity differences. For example,  \textsc{Para-Tgt} changes ``Man in a red shirt riding his bicycle'' to ``A bicyclist in a red shirt is riding''.
    Further, LLaMA transforms ``man on skis'' into ``skier'', ``person in blue and red ice climbing'' into ``ice climber'', and ``men with children'' into ``family''. The model tends to simplify, irrespective of the reference. For example, ``Two young people are approached by a flamboyant young woman dressed in a red bikini and a red feathered headress'' becomes ``Two young people are approached by a bikini-clad woman''. Paraphrasing could thus result in over-simplification.
    %or adding incorrect but common details. 
    %It could also result in adding incorrect but common details to captions; for example, Llama reasons that ``'Day' is often referred to as 'sunny day', 'beautiful day', or 'nice day'.'' However, in practice we find that targeted LlaMA paraphrasing based on reference sentences improves results. 

    \noindent \textbf{Human evaluation.} We extend quantification past retrieval by asking two German speakers to gauge the likelihood that captions are made by a German speaker and their naturalness. We provide 50 random captions for each of 3 sets (\textsc{Eng2Ger-MT}, 
    \textsc{Eng2Ger-HT},
    \textsc{Para-Tgt}).
    %\textit{Human Translation}). 
    Speakers do not know each set's identity and are tasked with scoring captions as 3=great, 2=good, 1=bad. On average, the speakers rate 
    \textsc{Eng2Ger-HT}
    %\textit{Human Translation} 
    the highest with a mean ternary score of 2.73 and mean binary score (great/good=1, bad=0) of 0.97. For \textsc{Para-Tgt}, the ternary score is 2.19 and binary score is 0.79. For \textsc{Eng2Ger-MT}, the ternary score is 2.16 and binary score is 0.77. These differences approximately reflect the recall results in Table \ref{tab:main}.

    \noindent \textbf{Recognition.} To evaluate object recognition, we compare objects mentioned in a native German caption to ones predicted by the models \textsc{Ger} and \textsc{Eng2Ger-HT}. We take predictions to be ones with CLIP scores greater than a threshold (the one in range 10:5:50 that maximizes val F1). A prediction is correct only if the object is mentioned in native German. Table \ref{tab:recognition_stats} shows train-set mentions and performance for the best-performing COCO supercategories.
    We observe large differences in the number of mentions, precision, and recall for several supercategories.  \textsc{Ger} achieves better recall (slightly correlated with mention count differences), but \textsc{Eng2Ger-HT} better precision. These results suggest potential recognition differences when using translated and native captions. 

    \begin{table}[t]
    \centering
    \renewcommand{\arraystretch}{0.88}
    \resizebox{\linewidth}{!}{
    \begin{tabular}{|c|c|c|c|c|c|}
    \hline
    
    \textbf{Supercategory} & \textbf{Vehicle}	& \textbf{Animal}	& \textbf{Sports}	& \textbf{Furniture}	& \textbf{Electronic}	\\ \hline
    \hline
    \textsc{Ger} (\#men) & 
    2604 &	2836 &	2101 &	\textbf{1488} &	510 \\
    \textsc{Eng2Ger-HT} (\#men) & 
    \textbf{2724}	& \textbf{2918}	& \textbf{2127}	& 1191	& \textbf{554}	 \\ \hline

    \textsc{Ger} (prec)    & 0.42	& 0.41	& 0.16		& 0.26	& 0.25	 \\ 
										
    \textsc{Eng2Ger-HT} (prec) & \textbf{0.47}	& \textbf{0.51}	& \textbf{0.17}	& \textbf{0.29}	& \textbf{0.27}	 \\ \hline

   \textsc{Ger} (rec) & \textbf{0.52}	& \textbf{0.55}	& \textbf{0.61}	& \textbf{0.20}	& 0.28	 \\
										
    \textsc{Eng2Ger-HT} (rec) & 0.46	& 0.44	& 0.56		& 0.16	& \textbf{0.30}	 \\ \hline

    \end{tabular} }
    \caption{\textbf{Recognition stats by supercategory}. Top rows: mention counts, middle: precision, bottom: recall.}
    \label{tab:recognition_stats}
\end{table}

% \begin{table*}[t]
%     \centering

%     \resizebox{\textwidth}{!}{
%     \begin{tabular}{|c|c|c|c|c|c|c|c|c|c|c|c|}
%     \hline
%     Supercat & Vehicle	& Outdoor	& Animal	& Accessory	& Sports	& Kitchen	& Food	& Furniture	& Electronic	& Appliance	& Indoor \\ \hline

%     Ger (mentions) & 
%     2604 &	\textbf{447} &	2836 &	477 &	2101 &	\textbf{463} & 135 &	\textbf{1488} &	510 &	34 &	195 \\
%     Ger-Indir (mentions) & 
%     \textbf{2724}	& 386	& \textbf{2918}	& \textbf{508}	& \textbf{2127}	& 359	& \textbf{183}	& 1191	& \textbf{554}	& \textbf{37}	& \textbf{214} \\ \hline

%     Ger (prec)    & 0.42	& 0.09	& 0.41	& \textbf{0.22}	& 0.16	& 0.10	& 0.15	& 0.26	& 0.25	& 0.05	& \textbf{0.09} \\ 
										
%     Ger-Indir (prec) & \textbf{0.47}	& \textbf{0.12}	& \textbf{0.51}	& 0.20	& \textbf{0.17}	& \textbf{0.13}	& \textbf{0.28}	& \textbf{0.29}	& \textbf{0.27}	& \textbf{0.07}	& \textbf{0.09} \\ \hline

%     Ger (rec) & \textbf{0.52}	& \textbf{0.17}	& \textbf{0.55}	& \textbf{0.19}	& \textbf{0.61}	& \textbf{0.10}	& \textbf{0.23}	& \textbf{0.20}	& 0.28	& \textbf{0.34}	& 0.18 \\
										
%     Ger-Indir (rec) & 0.46	& 0.14	& 0.44	& 0.14	& 0.56	& 0.06	& 0.17	& 0.16	& \textbf{0.30}	& \textbf{0.34}	& \textbf{0.34} \\ \hline

%     \end{tabular} }
%     \caption{Recognition stats by supercategory; top two rows = mention counts, middle two = precision, bottom two = recall}
%     \label{tab:recognition_stats}
% \end{table*}

\section{Conclusion}
    
We show notable differences in using native vs. translated German captions to train a retrieval model, and experiment with three strategies to reduce the gaps. We plan to extend investigation to more languages. Future work can also involve creation of data augmentation strategies that take inspiration from psychology literature \cite{nisbett2013culture,boroditsky2006linguistic} and solutions for the ambiguity challenges of machine translation, such as by using images \cite{futeral-etal-2023-tackling}.

\noindent \textbf{Acknowledgement.} This work was supported by NSF Grants No. 2006885 and 2329992.

\section*{Limitations and Ethical Considerations}

% THIS SECTION DOES NOT COUNT TOWARDS THE PAGE LIMIT

We only experiment with one translation model, one non-English language (German), and a small amount of runs of LLaMA-3. To ensure that insights generalize, various models, and languages (especially low-resource ones), should be analyzed. There may be intra-language variance amongst native speakers that should also be considered.

We rely on the use of image-caption datasets like Flickr30K and Multi30K. These datasets are relatively small (about 30k samples), so the coverage of concepts may not be fully representative of spoken language. Such datasets have also been noted to contain harmful biases with respect to attributes like race and gender \citep{van2016stereotyping}. The use of models like LLaMA-3 carries similar biases. There should be careful consideration regarding downstream usage of these sets and models. We note that a future extension of our paraphrasing strategies could be to mitigate the impact of in-group perspectives in the captions used for pretraining models. 

Our analysis of differences in languages is limited by the fact that languages are machine-translated to English. It is possible that some differences are amplified and/or missed due to machine translation artifacts.  

Finally, while we conduct initial human evaluation, we encourage larger-scale human evaluation that expands past our limited evaluation. This can be done to ensure that methods are applicable for a greater amount of people.

%\section{Ethical Considerations}
%    \input{latex/sections/ethical_cons}

\bibliography{custom}

\begin{thebibliography}{28}
\providecommand{\natexlab}[1]{#1}

\bibitem[{Berthele et~al.(2015)Berthele, Whelpton, N{\ae}ss, and Duijff}]{berthele2015static}
Raphael Berthele, Matthew Whelpton, {\AA}shild N{\ae}ss, and Pieter Duijff. 2015.
\newblock Static spatial descriptions in five {G}ermanic languages.
\newblock \emph{Language Sciences}, 49:82--101.

\bibitem[{Boroditsky(2006)}]{boroditsky2006linguistic}
Lera Boroditsky. 2006.
\newblock Linguistic relativity.
\newblock \emph{Encyclopedia of Cognitive Science}.

\bibitem[{Carlsson et~al.(2022)Carlsson, Eisen, Rekathati, and Sahlgren}]{carlsson2022cross-m-clip}
Fredrik Carlsson, Philipp Eisen, Faton Rekathati, and Magnus Sahlgren. 2022.
\newblock Cross-lingual and multilingual {CLIP}.
\newblock In \emph{Proceedings of the Thirteenth Language Resources and Evaluation Conference}, pages 6848--6854.

\bibitem[{Chen et~al.(2023)Chen, Hou, Chen, Dai, Shang, Jiang, Liu, Pan, and Wang}]{chen-etal-2023-mclip}
Guanhua Chen, Lu~Hou, Yun Chen, Wenliang Dai, Lifeng Shang, Xin Jiang, Qun Liu, Jia Pan, and Wenping Wang. 2023.
\newblock \href {https://doi.org/10.18653/v1/2023.acl-long.728} {m{CLIP}: Multilingual {CLIP} via cross-lingual transfer}.
\newblock In \emph{Proceedings of the 61st Annual Meeting of the Association for Computational Linguistics (Volume 1: Long Papers)}, pages 13028--13043, Toronto, Canada. Association for Computational Linguistics.

\bibitem[{Chen et~al.(2024)Chen, Djolonga, Padlewski, Mustafa, Changpinyo, Wu, Ruiz, Goodman, Wang, Tay et~al.}]{chen2023pali}
Xi~Chen, Josip Djolonga, Piotr Padlewski, Basil Mustafa, Soravit Changpinyo, Jialin Wu, Carlos~Riquelme Ruiz, Sebastian Goodman, Xiao Wang, Yi~Tay, et~al. 2024.
\newblock On scaling up a multilingual vision and language model.
\newblock In \emph{Proceedings of the IEEE/CVF Conference on Computer Vision and Pattern Recognition}, pages 14432--14444.

\bibitem[{Chen et~al.(2022)Chen, Wang, Changpinyo, Piergiovanni, Padlewski, Salz, Goodman, Grycner, Mustafa, Beyer et~al.}]{chen2022pali}
Xi~Chen, Xiao Wang, Soravit Changpinyo, AJ~Piergiovanni, Piotr Padlewski, Daniel Salz, Sebastian Goodman, Adam Grycner, Basil Mustafa, Lucas Beyer, et~al. 2022.
\newblock Pali: A jointly-scaled multilingual language-image model.
\newblock In \emph{The Eleventh International Conference on Learning Representations}.

\bibitem[{Conneau et~al.(2020)Conneau, Khandelwal, Goyal, Chaudhary, Wenzek, Guzm{\'a}n, Grave, Ott, Zettlemoyer, and Stoyanov}]{conneau-etal-2020-unsupervised}
Alexis Conneau, Kartikay Khandelwal, Naman Goyal, Vishrav Chaudhary, Guillaume Wenzek, Francisco Guzm{\'a}n, Edouard Grave, Myle Ott, Luke Zettlemoyer, and Veselin Stoyanov. 2020.
\newblock \href {https://doi.org/10.18653/v1/2020.acl-main.747} {Unsupervised cross-lingual representation learning at scale}.
\newblock In \emph{Proceedings of the 58th Annual Meeting of the Association for Computational Linguistics}, pages 8440--8451, Online. Association for Computational Linguistics.

\bibitem[{Elliott et~al.(2016)Elliott, Frank, Sima{'}an, and Specia}]{elliott-etal-2016-multi30k}
Desmond Elliott, Stella Frank, Khalil Sima{'}an, and Lucia Specia. 2016.
\newblock \href {https://doi.org/10.18653/v1/W16-3210} {{M}ulti30{K}: Multilingual {E}nglish-{G}erman image descriptions}.
\newblock In \emph{Proceedings of the 5th Workshop on Vision and Language}, pages 70--74, Berlin, Germany. Association for Computational Linguistics.

\bibitem[{Fan et~al.(2024)Fan, Krishnan, Isola, Katabi, and Tian}]{fan2024improving}
Lijie Fan, Dilip Krishnan, Phillip Isola, Dina Katabi, and Yonglong Tian. 2024.
\newblock Improving clip training with language rewrites.
\newblock \emph{Advances in Neural Information Processing Systems}, 36.

\bibitem[{Futeral et~al.(2023)Futeral, Schmid, Laptev, Sagot, and Bawden}]{futeral-etal-2023-tackling}
Matthieu Futeral, Cordelia Schmid, Ivan Laptev, Beno{\^\i}t Sagot, and Rachel Bawden. 2023.
\newblock \href {https://doi.org/10.18653/v1/2023.acl-long.295} {Tackling ambiguity with images: Improved multimodal machine translation and contrastive evaluation}.
\newblock In \emph{Proceedings of the 61st Annual Meeting of the Association for Computational Linguistics (Volume 1: Long Papers)}, pages 5394--5413, Toronto, Canada. Association for Computational Linguistics.

\bibitem[{Hodosh et~al.(2013)Hodosh, Young, and Hockenmaier}]{hodosh2013framing}
Micah Hodosh, Peter Young, and Julia Hockenmaier. 2013.
\newblock Framing image description as a ranking task: Data, models and evaluation metrics.
\newblock \emph{Journal of Artificial Intelligence Research}, 47:853--899.

\bibitem[{Hofstede(2001)}]{hofstede2001culture}
Geert Hofstede. 2001.
\newblock Culture's consequences: Comparing values, behaviors, institutions and organizations across nations.
\newblock \emph{Thousand Oaks}.

\bibitem[{Hu et~al.(2019)Hu, Rudinger, Post, and Van~Durme}]{hu2019parabank}
J~Edward Hu, Rachel Rudinger, Matt Post, and Benjamin Van~Durme. 2019.
\newblock Parabank: Monolingual bitext generation and sentential paraphrasing via lexically-constrained neural machine translation.
\newblock In \emph{Proceedings of the AAAI Conference on Artificial Intelligence}, volume~33, pages 6521--6528.

\bibitem[{K{\'a}d{\'a}r et~al.(2018)K{\'a}d{\'a}r, Elliott, C{\^o}t{\'e}, Chrupa{\l}a, and Alishahi}]{kadar2018lessons}
{\'A}kos K{\'a}d{\'a}r, Desmond Elliott, Marc-Alexandre C{\^o}t{\'e}, Grzegorz Chrupa{\l}a, and Afra Alishahi. 2018.
\newblock Lessons learned in multilingual grounded language learning.
\newblock In \emph{Proceedings of the 22nd Conference on Computational Natural Language Learning}, pages 402--412.

\bibitem[{Kay and Kempton(1984)}]{kay1984sapir}
Paul Kay and Willett Kempton. 1984.
\newblock What is the {S}apir-{W}horf hypothesis?
\newblock \emph{American Anthropologist}, 86(1):65--79.

\bibitem[{Lin et~al.(2014)Lin, Maire, Belongie, Hays, Perona, Ramanan, Doll{\'a}r, and Zitnick}]{lin2014microsoft}
Tsung-Yi Lin, Michael Maire, Serge Belongie, James Hays, Pietro Perona, Deva Ramanan, Piotr Doll{\'a}r, and C~Lawrence Zitnick. 2014.
\newblock Microsoft {COCO}: Common objects in context.
\newblock In \emph{Computer Vision--ECCV 2014: 13th European Conference, Zurich, Switzerland, September 6-12, 2014, Proceedings, Part V 13}, pages 740--755. Springer.

\bibitem[{Liu et~al.(2021)Liu, Bugliarello, Ponti, Reddy, Collier, and Elliott}]{liu2021visually}
Fangyu Liu, Emanuele Bugliarello, Edoardo~Maria Ponti, Siva Reddy, Nigel Collier, and Desmond Elliott. 2021.
\newblock Visually grounded reasoning across languages and cultures.
\newblock \emph{Empirical Methods In Natural Language Processing}.

\bibitem[{Lu et~al.(2018)Lu, Yang, Batra, and Parikh}]{lu2018neural}
Jiasen Lu, Jianwei Yang, Dhruv Batra, and Devi Parikh. 2018.
\newblock Neural baby talk.
\newblock In \emph{Proceedings of the IEEE Conference on Computer Vision and Pattern Recognition}, pages 7219--7228.

\bibitem[{Miller(1995)}]{miller1995wordnet}
George~A Miller. 1995.
\newblock Word{N}et: A lexical database for {E}nglish.
\newblock \emph{Communications of the ACM}, 38(11):39--41.

\bibitem[{Nisbett and Masuda(2013)}]{nisbett2013culture}
Richard~E Nisbett and Takahiko Masuda. 2013.
\newblock Culture and point of view.
\newblock In \emph{Biological and {C}ultural {B}ases of {H}uman {I}nference}, pages 49--70. Psychology Press.

\bibitem[{Radford et~al.(2021)Radford, Kim, Hallacy, Ramesh, Goh, Agarwal, Sastry, Askell, Mishkin, Clark et~al.}]{radford2021learning}
Alec Radford, Jong~Wook Kim, Chris Hallacy, Aditya Ramesh, Gabriel Goh, Sandhini Agarwal, Girish Sastry, Amanda Askell, Pamela Mishkin, Jack Clark, et~al. 2021.
\newblock Learning transferable visual models from natural language supervision.
\newblock In \emph{International Conference on Machine Learning}, pages 8748--8763. PMLR.

\bibitem[{Thapliyal et~al.(2022)Thapliyal, Pont-Tuset, Chen, and Soricut}]{ThapliyalCrossmodal2022}
Ashish Thapliyal, Jordi Pont-Tuset, Xi~Chen, and Radu Soricut. 2022.
\newblock {Crossmodal-3600: A Massively Multilingual Multimodal Evaluation Dataset}.
\newblock In \emph{EMNLP}.

\bibitem[{Tiedemann and Thottingal(2020)}]{TiedemannThottingal:EAMT2020}
J{\"o}rg Tiedemann and Santhosh Thottingal. 2020.
\newblock {OPUS-MT} — {B}uilding open translation services for the {W}orld.
\newblock In \emph{Proceedings of the 22nd Annual Conferenec of the European Association for Machine Translation (EAMT)}, Lisbon, Portugal.

\bibitem[{Touvron et~al.(2023)Touvron, Lavril, Izacard, Martinet, Lachaux, Lacroix, Rozi{\`e}re, Goyal, Hambro, Azhar et~al.}]{touvron2023llama}
Hugo Touvron, Thibaut Lavril, Gautier Izacard, Xavier Martinet, Marie-Anne Lachaux, Timoth{\'e}e Lacroix, Baptiste Rozi{\`e}re, Naman Goyal, Eric Hambro, Faisal Azhar, et~al. 2023.
\newblock Llama: Open and efficient foundation language models.
\newblock \emph{arXiv preprint arXiv:2302.13971}.

\bibitem[{Van~Miltenburg(2016)}]{van2016stereotyping}
Emiel Van~Miltenburg. 2016.
\newblock Stereotyping and bias in the {F}lickr30k dataset.
\newblock \emph{Proceedings of the Workshop on Multimodal Corpora (MMC)}.

\bibitem[{Wieting and Gimpel(2018)}]{wieting-gimpel-2018-paranmt}
John Wieting and Kevin Gimpel. 2018.
\newblock \href {https://doi.org/10.18653/v1/P18-1042} {{P}ara{NMT}-50{M}: Pushing the limits of paraphrastic sentence embeddings with millions of machine translations}.
\newblock In \emph{Proceedings of the 56th Annual Meeting of the Association for Computational Linguistics (Volume 1: Long Papers)}, pages 451--462, Melbourne, Australia. Association for Computational Linguistics.

\bibitem[{Yoshikawa et~al.(2017)Yoshikawa, Shigeto, and Takeuchi}]{yoshikawa-etal-2017-stair}
Yuya Yoshikawa, Yutaro Shigeto, and Akikazu Takeuchi. 2017.
\newblock \href {https://doi.org/10.18653/v1/P17-2066} {{STAIR} captions: Constructing a large-scale {J}apanese image caption dataset}.
\newblock In \emph{Proceedings of the 55th Annual Meeting of the Association for Computational Linguistics (Volume 2: Short Papers)}, pages 417--421, Vancouver, Canada. Association for Computational Linguistics.

\bibitem[{Young et~al.(2014)Young, Lai, Hodosh, and Hockenmaier}]{young-etal-2014-image}
Peter Young, Alice Lai, Micah Hodosh, and Julia Hockenmaier. 2014.
\newblock \href {https://doi.org/10.1162/tacl_a_00166} {From image descriptions to visual denotations: New similarity metrics for semantic inference over event descriptions}.
\newblock \emph{Transactions of the Association for Computational Linguistics}, 2:67--78.

\end{thebibliography}

\appendix

\section*{Appendix}
\label{sec:appendix}

    Shown are the prompt templates used for querying LLaMA-3 (meta-llama/Meta-Llama-3-8B-Instruct on Hugging Face). We do not experiment with LLaMA sampling settings and generate outputs with default parameters.  

    \begin{quote}
        \textit{Para-Rnd Prompt Template} \\
        Rewrite captions in a structurally different manner, while closely maintaining semantic meaning. Return as Python string. Return no other text. 
    \end{quote}
    
    \begin{quote}
        \textit{Para-Tgt Prompt Template} \\
        1) Given a caption, 1st decompose into noun phrases, keeping all phrase content (e.g. adjectives) aside from articles. 
        EX: ``A person is riding a blue bicycle down the street on a sunny day.''
        Noun Phrases: [``person'', ``blue bicycle'', ``street'', ``sunny day'']
        
        2) Based on a provided reference list of related captions, construct a new set of noun phrases that alters the original noun phrases to be in the common styles/forms shown in the reference list.
        EX: If many captions say ``bicyclist'', combine ``person'' and ``blue bicycle'' into ``bicyclist''.
        Do not infer unnecessary information. 
        
        3) Finally, combine the new noun phrases back into a sentence, keeping the same semantics as the original caption.
        EX: ``A bicyclist is traveling down the road on a sunny day.'' 
        
        Here is your reference caption list: \{ref$_{caps}$\}
        
        Now run each steps 1-3 for the example: 
        ``\{example\}'' 
        
        Enclose the final output caption in <final></final> tags for easy parsing.  
    
    \end{quote}

    \begin{quote}
        \textit{System Prompt for Experiments} \\
        I'm a researcher using LLMs for NLP tasks. Behave like an automatic processing agent for the user.
    \end{quote}

\end{document}